\def\year{2022}\relax
\documentclass[letterpaper]{article} 
\usepackage{aaai22}  
\usepackage{times}  
\usepackage{helvet}  
\usepackage{courier}  
\usepackage[hyphens]{url}  
\usepackage{graphicx} 
\def\UrlFont{\rm}  
\usepackage{natbib}  
\usepackage{caption} 
\DeclareCaptionStyle{ruled}{labelfont=normalfont,labelsep=colon,strut=off} 
\usepackage[titlenumbered,ruled]{algorithm2e}
\usepackage{algorithmicx}

\usepackage{xcolor}
\usepackage{multirow}
\usepackage{booktabs}
\usepackage{nicefrac}

\usepackage{amsmath}
\usepackage{amsfonts}
\makeatletter
\newcommand{\removelatexerror}{\let\@latex@error\@gobble}
\makeatother
\usepackage{newfloat}
\title{Early-Bird GCNs: Graph-Network Co-Optimization Towards More Efficient GCN Training and Inference via Drawing Early-Bird Lottery Tickets}

\author{
    Haoran You, 
    Zhihan Lu,
    Zijian Zhou,
    Yonggan Fu,
    Yingyan Lin
}
\affiliations{
    Department of Electrical and Computer Engineering, Rice University\\

    \{haoran.you, zl55, zjzhou, yf22, yingyan.lin\}@rice.edu
}

\usepackage{bibentry}

\begin{document}

\maketitle

\begin{abstract}
Graph Convolutional Networks (GCNs) have emerged as the state-of-the-art deep learning model for representation learning on graphs. However, it remains notoriously challenging to train and inference GCNs over large graph datasets, limiting their application to large real-world graphs and hindering the exploration of deeper and more sophisticated GCN graphs. This is because as the graph size grows, the sheer number of node features and the large adjacency matrix can easily explode the required memory and data movements. To tackle the aforementioned challenges, we explore the possibility of drawing lottery tickets when sparsifying GCN graphs, i.e., subgraphs that largely shrink the adjacency matrix yet are capable of achieving accuracy comparable to or even better than their full graphs. Specifically, we for the first time discover the existence of graph early-bird (GEB) tickets that emerge at the very early stage when sparsifying GCN graphs, and propose a simple yet effective detector to automatically identify the emergence of such GEB tickets. Furthermore, we advocate graph-model co-optimization and develop a generic efficient GCN early-bird training framework dubbed GEBT that can significantly boost the efficiency of GCN training by (1) drawing joint early-bird tickets between the GCN graphs and models and (2) enabling simultaneously sparsification of both the GCN graphs and models.
Experiments on various GCN models and datasets consistently validate our GEB finding and the effectiveness of our GEBT, e.g., 
our GEBT achieves up to 80.2\% $\sim$ 85.6\% and 84.6\% $\sim$ 87.5\% savings of GCN training and inference costs while offering a comparable or even better accuracy as compared to state-of-the-art methods. 
Our source code and supplementary material are available at \textcolor{blue}{\url{https://github.com/RICE-EIC/Early-Bird-GCN}}.

\end{abstract}

\vspace{-1em}
\section{Introduction}
\vspace{-0.2em}

Graph convolutional networks (GCNs) \cite{kipf2016semi} have emerged as state-of-the-art (SOTA) algorithms for graph-based learning tasks, such as graph classification \cite{xu2018powerful} and node classification \cite{kipf2016semi}.
It is well recognized that the superior performance
largely benefits from GCNs' ability for handling irregularity and unrestricted neighborhood connections.
Specifically, for each node in a graph, GCNs first aggregate neighbor nodes' features, and then transform the aggregated feature through (hierarchical) feed-forward propagation to update the feature of the given node. 

Despite their promise, GCN training and inference can be notoriously challenging, hindering their great potential from being unfolded in large real-world graphs. This is because as the graph dataset grows, the large number of node features and the abundant adjacency matrix can easily explode the required memory and data movements \cite{10.1145/3466752.3480113,DBLP:journals/corr/abs-2109-08983}. For example, a mere 2-layer GCN model with 32 hidden units requires 19 GFLOPs (FLOPs: floating point operations) to process the Reddit graph \cite{tailor2020degree}, twice as much as that of a powerful deep neural network (DNN) ResNet50, which has a total of 8 GFLOPs when processing ImageNet \cite{canziani2016analysis}. The giant computational cost of GCNs comes from three aspects. \underline{First}, graphs (or graph data), especially real-world ones, are often extraordinarily large and irregular as exacerbated by their intertwined complex neighbor connections, e.g., 
a total of 232,965 nodes in the Reddit graph with each node having about 50 neighbors \cite{KKMMN2016}. \underline{Second}, the dimension of GCNs' node feature vectors can be very high, e.g., each node in the Citeseer graph has 3703 features. \underline{Third}, the extremely high sparsity and unbalanced
distribution of 
non-zero data in GCNs' adjacency matrices imposes a paramount challenge for effectively accelerating GCNs \cite{geng2020awb,yan2020hygcn}, e.g., as high as 99.9\% vs. 10\% to 50\% generally observed in DNNs.


To tackle the aforementioned challenges and unleash the full potential of GCNs, various techniques have been developed. For instance, Tailor et al. \cite{tailor2020degree} leverages quantization-aware training to demonstrate 8-bit GCNs; SGCN \cite{li2020sgcn} is the first to consider GCN sparsification by formulating and solving it as an optimization problem.


The impressive performance achieved by existing GCN compression works indicates that there are redundancies within GCNs to be leveraged for aggressively trimming down their complexity while maintaining their performance. In this work, we attempt to take a new perspective by drawing inspiration from the tremendous success of DNN compression, particularly the lottery ticket (LT) finding \cite{frankle2018the,liu2018rethinking,You2020Drawing}. 
While conceptually simple, the unique structures of GCNs make it not straightforward to leverage the LT finding to compress GCNs.
This is because (1) the graph instead of the MLPs in GCNs dominates the complexity, for which the existence of LT remains unknown; and (2) it is unclear how to jointly optimize the two phases of GCN operations (i.e., feature \textit{aggregation} and \textit{combination}) while doing so promises the maximum complexity reduction. 

This paper aims to close the above gap to minimize the complexity of GCNs without hurting their competitive performance, and to make the following contributions: 

\begin{itemize}
\itemsep -0.3\parsep
    \item We discover the existence of graph early-bird (GEB) tickets that emerge at the very early stage when sparsifying GCN graphs, and propose a simple yet effective detector to automatically identify the emergence of GEB tickets. To our best knowledge, we are the first to show that the 
    early-bird tickets
    finding holds for GCN graphs.  

    \item we advocate graph-network co-optimization and develop a generic efficient GCN training framework dubbed GEBT that significantly boosts GCN training efficiency by (1) drawing joint early-bird (EB) tickets between the GCN graphs and models and (2) simultaneously sparsifying both the GCN graphs and models, additionally boosting the GCN inference efficiency.
    
    
    \item Experiments on various GCN models and datasets consistently validate our GEB finding and the effectiveness of the proposed GEBT. For example, our GEBT achieves up to 80.2\% $\sim$ 85.6\% and 84.6\% $\sim$ 87.5\% GCN training and inference costs savings while leading to a comparable or even better accuracy as compared to state-of-the-art (SOTA) methods. 
    
\end{itemize}

\section{Related Works}

\begin{figure*}[t]
    \centering
    \includegraphics[width=0.85\linewidth]{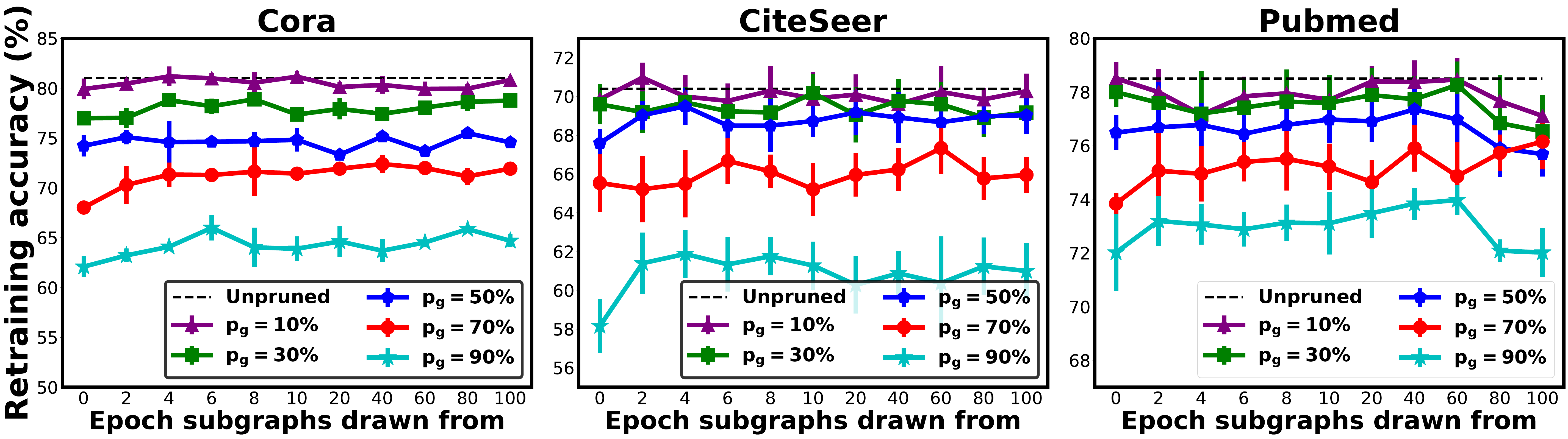}
      \vspace{-0.5em}
    \caption{Retraining accuracy vs. epoch numbers at which subgraphs are drawn, when evaluating the GCNs \cite{kipf2017semi} on three graph datasets: Cora, Citeseer, and Pumbed, where dashed lines show the accuracy of GCNs on corresponding unpruned full graphs, $p_g$ denotes the graph pruning ratios, and error bars show the minimum and maximum of ten runs.}
    \vspace{-1em}
    \label{fig:GEB_tickets}
\end{figure*}

\textbf{Graph Convolutional Networks (GCNs).}
GCNs have amazed us for processing non-Euclidean and irregular data structures \cite{zhang2018end}. Recently developed GCNs can be categorized into two groups: spectral and spatial methods. Specifically, spectral methods \cite{kipf2017semi, peng2020learning} model the representation in the graph Fourier transform domain based on eigen-decomposition, which are time-consuming and usually handle the whole graph simultaneously making it difficult to parallel or scale to large graphs \cite{gao2019graphnas, wu2020comprehensive}. On the other hand, spatial approaches \cite{hamilton2017inductive,simonovsky2017dynamic}, which directly perform the convolution in the graph domain by aggregating the neighbor nodes’ information, have rapidly developed recently. To further improve the performance of spatial GCNs, Veličković et al. \cite{GAT} introduce the attention mechanism to select information which is relatively critical from all inputs; Zeng et al. \cite{zeng2019accurate} propose mini-batch training to improve GCNs' scalability of handling large graphs; and \cite{xu2018how} theoretically formalizes an upper bound for the expressiveness of GCNs. Our GEB finding and GEBT enhance the understanding of GCNs and promote efficient GCN training, and can be generally applicable to SOTA GCN models.  

\textbf{GCN Compression.} The prohibitive complexity and powerful performance of GCNs have motivated growing interest in GCN compression.
For instance, Tailor et al. \cite{tailor2020degree} for the first time show the feasibility of adopting 8-bit integer arithmetic representation for GCN inference without sacrificing the classification accuracy; two concurrent pruning works \cite{li2020sgcn,zheng2020robust} aim to sparsify the graph adjacency matrices; and 
Ying et al. \cite{ying2018hierarchical} propose a DiffPool layer to reduce the size of GCN graphs by clustering similar nodes during training and inference. Our GEBT explores from a new perspective and is complementary with exiting GCN compression works, i.e., can be applied on top of them to further reduce GCNs' training/inference costs. 

\textbf{Early-Bird Tickets Hypothesis.}
Frankle et al. \cite{frankle2018the} show that winning tickets (i.e., small subnetworks) exist in randomly initialized dense networks, which can be retrained to restore a comparable or even better performance than their dense network counterparts. This finding has attracted lots of research attentions as it implies the potential of training a much smaller network to reach the accuracy of a dense, much larger network without going through the time and cost consuming pipeline of fully training the dense network, pruning and then retraining it to restore the accuracy. Later, You et al. \cite{You2020Drawing} demonstrate the existence of EB tickets, i.e., the winning tickets can be consistently drawn at the very early training stages across different models and datasets, and leverages this to largely reduce the training costs of DNNs. More recently, the EB finding has been extended to natural language processing (NLP) models (e.g., BERT) \cite{chen2021earlybert} and generative adversarial networks (GANs) \cite{mukund2020winning}. Our GEB finding and GEBT draw inspirations from the prior arts, and for the first time demonstrate that the EB phenomenon holds for GCNs which have unique and different algorithm structures as compared to DNNs, NLP, and GANs. 
Furthermore, compared with the iterative pruning method, e.g., UGS \cite{chen2021unified}, we \textbf{for the first time} show that \textit{early-bird (EB)} tickets exist in both GCN graphs and networks, and further develop efficient and effective detectors to automatically identify them, boosting both training and inference efficiency, while UGS draws \textit{lottery tickets} after \textit{fully and iteratively (up to 20$\times$)} training the dense models for only saving inference costs.

\section{Our Findings and Proposed Techniques}

\subsection{Preliminaries of GCNs and GCN Sparsification}
\textbf{GCN Notation and Formulation.} Let $G = (V, E)$ represents a GCN graph, where $v_i \in V$ and $(v_i, v_j) \in E$  denote the nodes and edges, respectively; and $N = | V |$ and $M = | E |$ denote the total number of nodes and edges, respectively. The node degrees are denoted as $d = \{d_1, d_2, \cdots, d_N\}$ where $d_i$ indicates the number of neighbors connected to the node $v_i$. We define $D$ as the degree matrix whose diagonal elements are formed using $d$. Given the adjacency matrix $A$ and the feature matrix $X = \{x_1, x_2, \cdots, x_N\}$ of the graph $G$, a two-layer GCN model \cite{kipf2017semi} can then be formulated as:
\begin{equation}\label{eq:gcn}
    Z \!=\! f(A, X) \!=\! \text{softmax} \left(\hat{A} \, \text{ReLU} \left(\hat{A}XW_0\right) W_1 \right),
\end{equation}
where $\hat{A} = D^{-\frac{1}{2}} (A + I_n) D^{-\frac{1}{2}}$ is calculated by a pre-processing step, thus multiplying $\hat{A}$ captures GCNs' neighbor aggregation; $W_0$ and $W_1$ are the weights of the GCN model for the 1st and 2nd layers, e.g., $W_0$ is an input-to-hidden weight matrix for a hidden layer with $H$ feature maps and $W_1$ is a hidden-to-output weight matrix with $F$ feature maps (i.e., $Z \in \mathbb{R}^{N \times F}$), where the mapping from the input to the hidden or output layer is called GCN combination which combines each node's features and its neighbors; The softmax function $\text{softmax} (x_i) = \text{exp} (x_i) / \sum_i \text{exp} (x_i)$ is applied in a row-wise manner \cite{kipf2017semi}. For semi-supervised multiclass classification, the loss function of the cross-entropy errors over all labeled examples:
\begin{equation}\label{eq:gcn_loss}
    \mathcal{L}_{GCN}(W) = - \sum_{n \in \mathcal{Y}_N} \sum_f Y_{nf} \, ln(Z_{nf}),
\end{equation}
where $\mathcal{Y}_N$ is the set of node indices that have labels, $Y_{nf}$ and $Z_{nf}$ are the ground truth label matrix and the GCN output predictions, respectively. During GCN training, $W_0$ and $W_1$ are updated. 
via gradient descents.

\textbf{Graph Sparsification.} The goal of graph sparsification is to reduce the total number of edges in GCNs' graph (i.e., the size of the adjacency matrices). A SOTA graph sparsification pipeline \cite{li2020sgcn} is to first pretrain GCNs on their full graphs, and then sparsify the graphs based on the pretrained GCNs. The weights of GCNs are not updated during graph sparsification, during which $W$ is replaced with $A$ in Eq. (\ref{eq:gcn_loss}) to derive the loss function $\mathcal{L}_{GCN}(A)$. The overall loss function during graph sparsification can be written as:
\begin{equation} \label{eq:graph_loss}
    \mathcal{L}_{Graph}(A) = \mathcal{L}_{GCN}(A) + \mathcal{L}_{Reg}(A),
\end{equation}
where $\mathcal{L}_{Reg}$ denotes the sparse regularization term, which ideally will become zero if the sparsity of the graph adjacency matrices reaches the specified pruning ratio (e.g., $\nicefrac{\|A_{prune}\|_0}{\|A\|_0} \leq 1 - p$ for a given ratio of $p$). As  $\mathcal{L}_{Reg}$ is not differentiable, SOTA graph sparsification work \cite{li2020sgcn} formulates Eq. (\ref{eq:graph_loss}) as an alternating optimization problem for updating the graph adjacency matrices.


\subsection{Finding 1: EB Tickets Exist in GCN Graphs}
\label{sec:GEB_tickets}

In this subsection, we first conduct an extensive set of experiments to show
that GEB tickets can be observed across popular graph datasets, and then propose a simple yet effective method to detect the emergence of GEB tickets. 

\textbf{Experiment Settings.} For this set of experiments, we follow the SOTA graph sparsification work \cite{li2020sgcn} to first \textit{pretrain} GCNs on unpruned graphs, \textit{train and prune} the graphs based on the pretrained GCNs, and then \textit{retrain} GCNs from scratch on the pruned graphs to evaluate the achieved accuracy. In addition, we adopt a two-layer GCN as described in Eq. (\ref{eq:gcn}), in which both the GCN and graph training take a total of 100 epochs and 
an Adam solver is used with a learning rate of 0.01 and 0.001 for training the GCNs and graphs, respectively. For retraining the pruned graphs, we keep the same setting by default.

\textbf{Existence of GEB Tickets.} 
We follow the SOTA method \cite{li2020sgcn} to sparsify the graph, but instead prune the graph that \textit{have not been fully trained} (before the accuracy reaches their final top values), to see if reliable GEB tickets can be observed, i.e., the retraining accuracy reaches the one drawn from the corresponding fully-trained graph. Fig. \ref{fig:GEB_tickets} shows the accuracies achieved by re-training the pruned graphs drawn from different early epochs, considering three different graph datasets and six pruning ratios. \textbf{Two intriguing observations} can be made: (1) there consistently exist GEB tickets drawn from certain early epochs (e.g., as early as 10 epochs w.r.t. the total of 100 epochs), of which the retraining accuracy is comparable or even better than those drawn in a later stage, including the ``ground-truth” tickets drawn from the fully-trained graphs (i.e., at the 100-th epoch); and (2) some GEB tickets (e.g., $P_g=30\%$ on Pumbed) can even outperform their unpruned graphs (denoted using dashed lines), potentially thanks to the sparse regularization as mentioned in \cite{You2020Drawing}. The first observation implies the possibility of ``overcooking” when identifying important graph edges at later training stages.

\begin{figure}
    \centering
    \includegraphics[width=\linewidth]{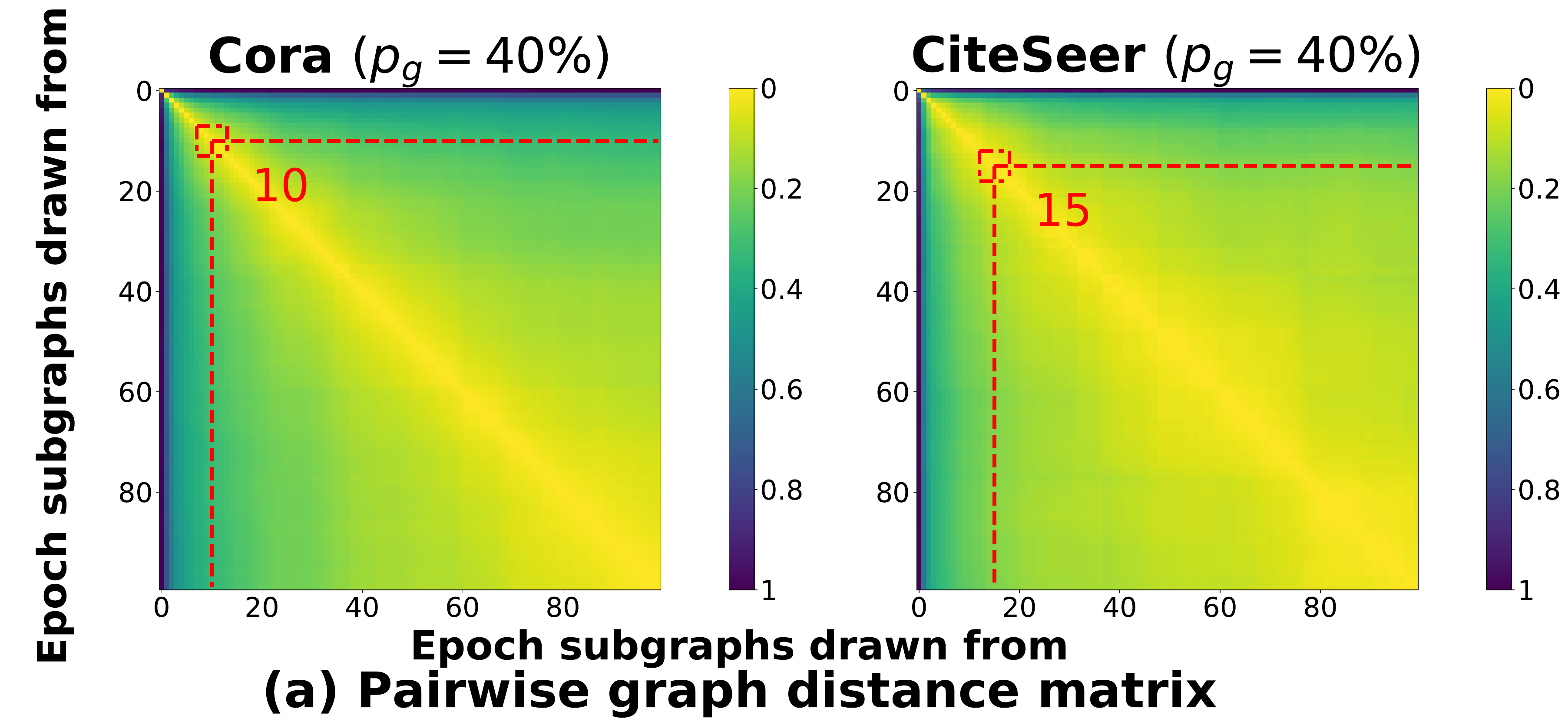}
    \includegraphics[width=0.95\linewidth]{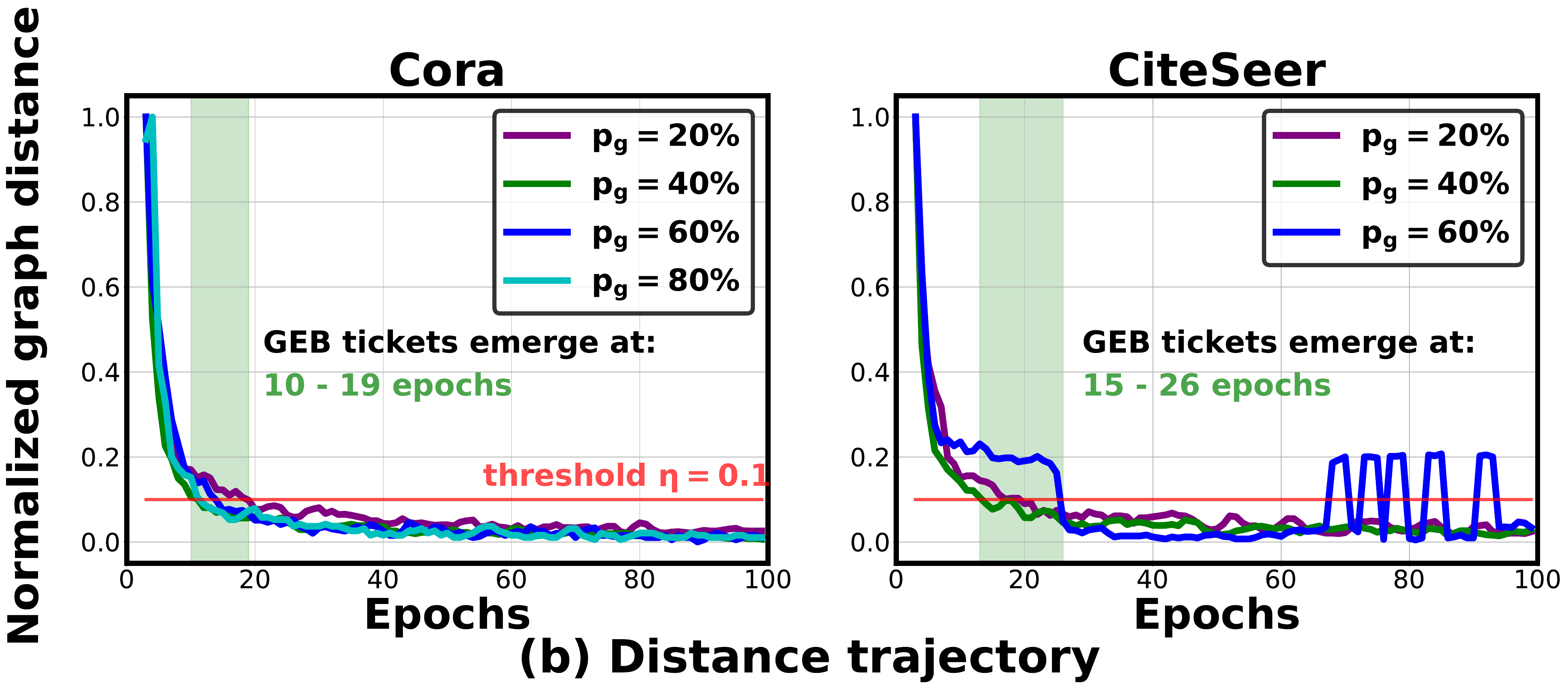}
    \vspace{-0.5em}
    \caption{The visualization of (a) pairwise graph distance matrices, and (b) recorded graph distance's evolution along the training trajectories under different graph pruning ratios.}
    \vspace{-1.5em}
    \label{fig:GEB_identification}
\end{figure}

\textbf{Detection of GEB Tickets.} The existence of GEB tickets and the prohibitive cost of GCN training motivate us to explore the possibility of automatically detecting the emergence of GEB tickets. To do so, we develop a simple yet effective detector via measuring the ``graph distance'' between consecutive epochs during graph sparsification. Specifically, we define a binary mask of the drawn GEB tickets (i.e., pruned graphs), where 1 denotes the reserved edges and 0 denotes the pruned edges, and use the hamming distance between the corresponding masks to measure the  ``distance'' between two graphs.

Fig. \ref{fig:GEB_identification} (a) visualizes the pairwise ``graph distance'' matrices among 100 training epochs, where the $(i, j)$-th element within the matrices represents the distance between the pruned graphs drawn at the $i$-th and $j$-th epochs. We see that the distance deceases rapidly (i.e., color change from green to yellow) at the first few epochs, indicating that the reserved edges in pruned graphs quickly converge at the very early training stages. We therefore measure and record the distance between consecutive three epochs (i.e., look back for three epochs during training), and stop training the graph when all the recorded distances are smaller than a specified threshold $\eta$. Fig. \ref{fig:GEB_identification} (b) plots the maximum recorded distances as graph training epochs increase, where the red line denotes the threshold we adopt in all experiments with different pruning ratios. The identified GEB tickets are consistently drawn from the early (10- $\sim$ 26-th) epochs. These experiments validate the effectiveness of our developed GEB detector, which has negligible overheads compared with the total graph training cost (i.e., $<$ 0.1\%).

\begin{figure}[t]
    \centering
    \includegraphics[width=0.9\linewidth]{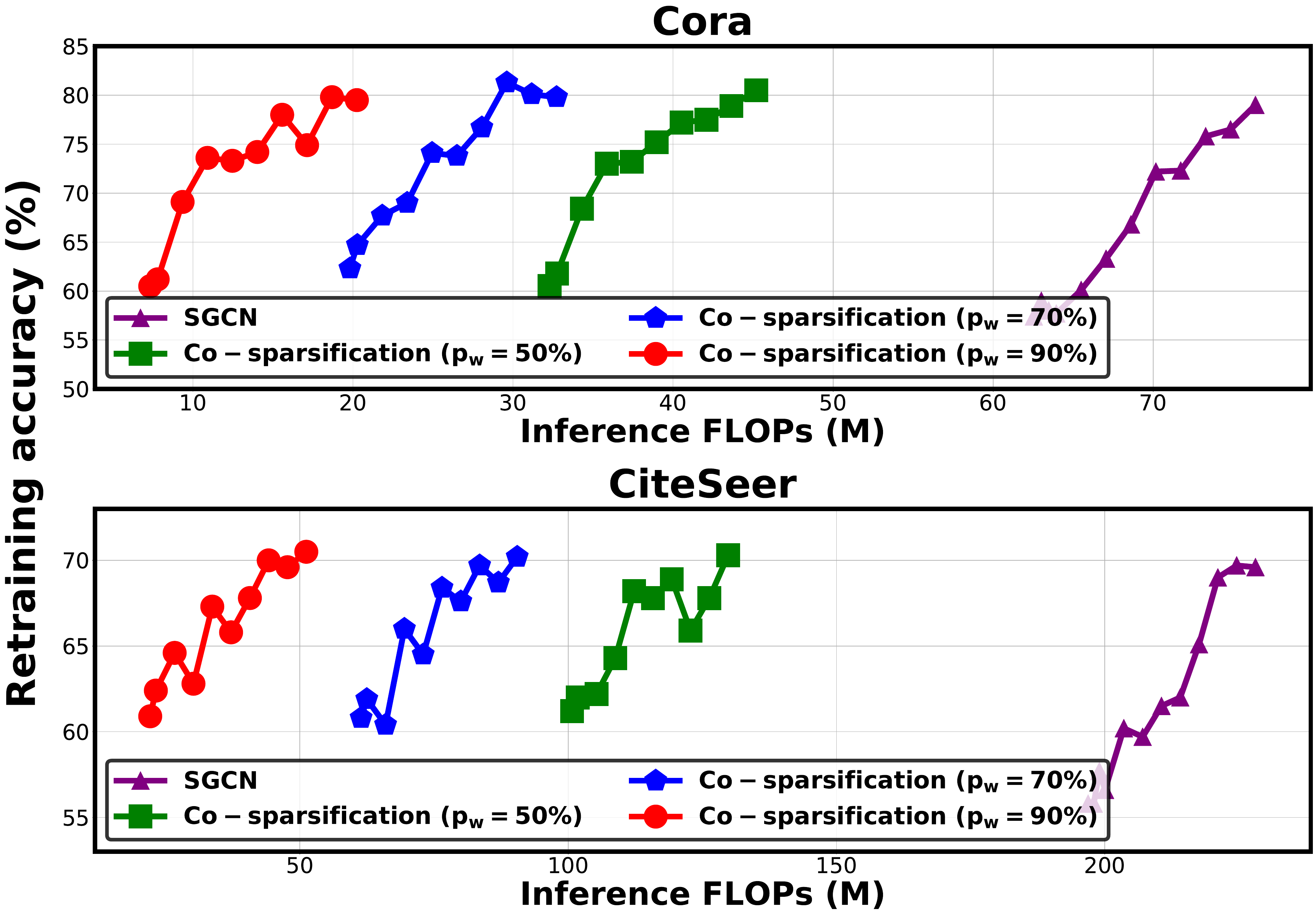}
     \vspace{-1em}
    \caption{Retraining accuracy vs. inference FLOPs of our co-sparsification framework and a SOTA graph sparsification framework, SGCN \cite{li2020sgcn}.}
    \vspace{-1.em}
    \label{fig:Co-optimization}
\end{figure}

\begin{figure*}[t]
    \centering
    \vspace{-0.5em}
    \includegraphics[width=0.72\linewidth,height=0.4\linewidth]{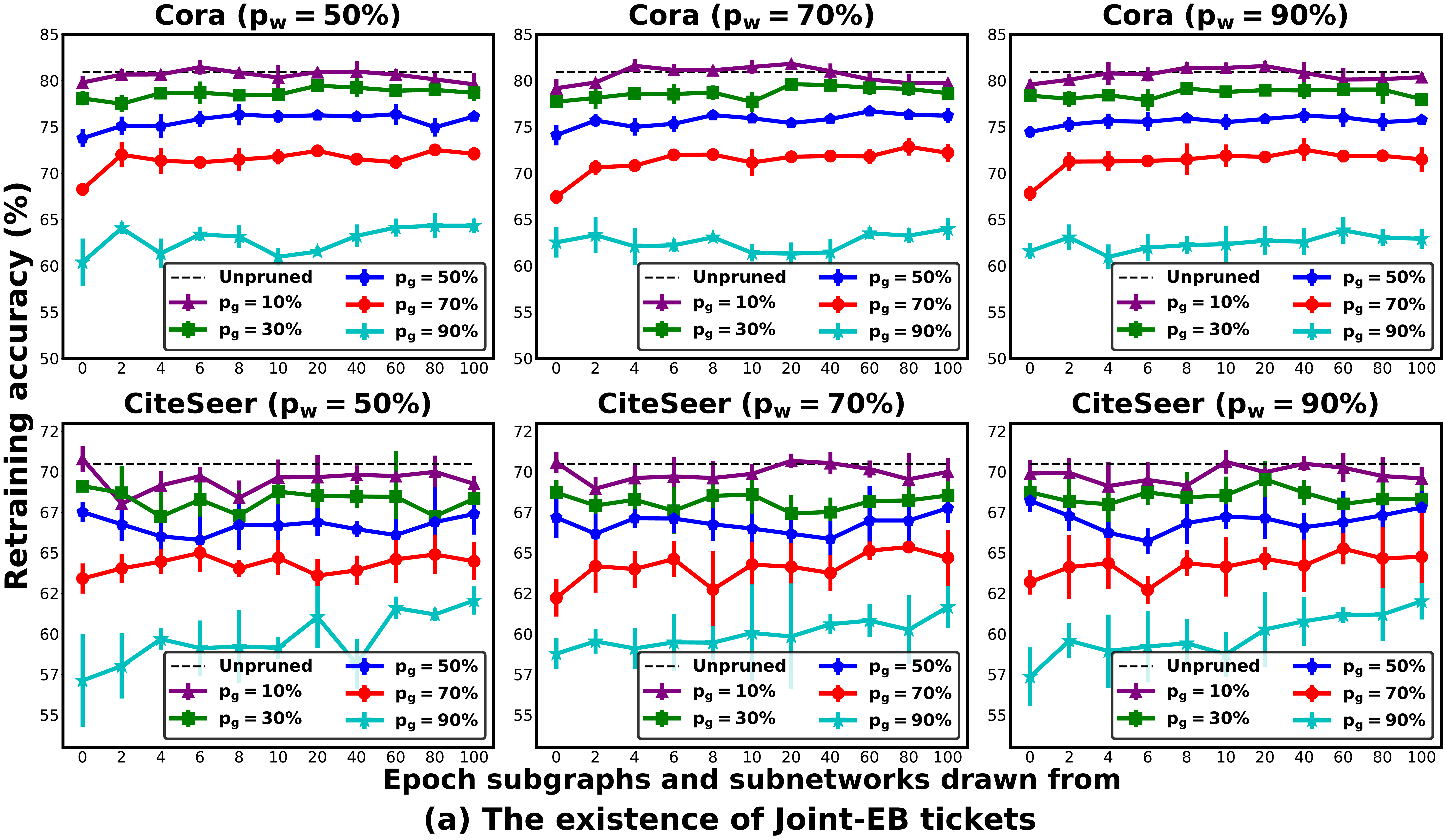}
    \includegraphics[width=0.26\linewidth,height=0.4\linewidth]{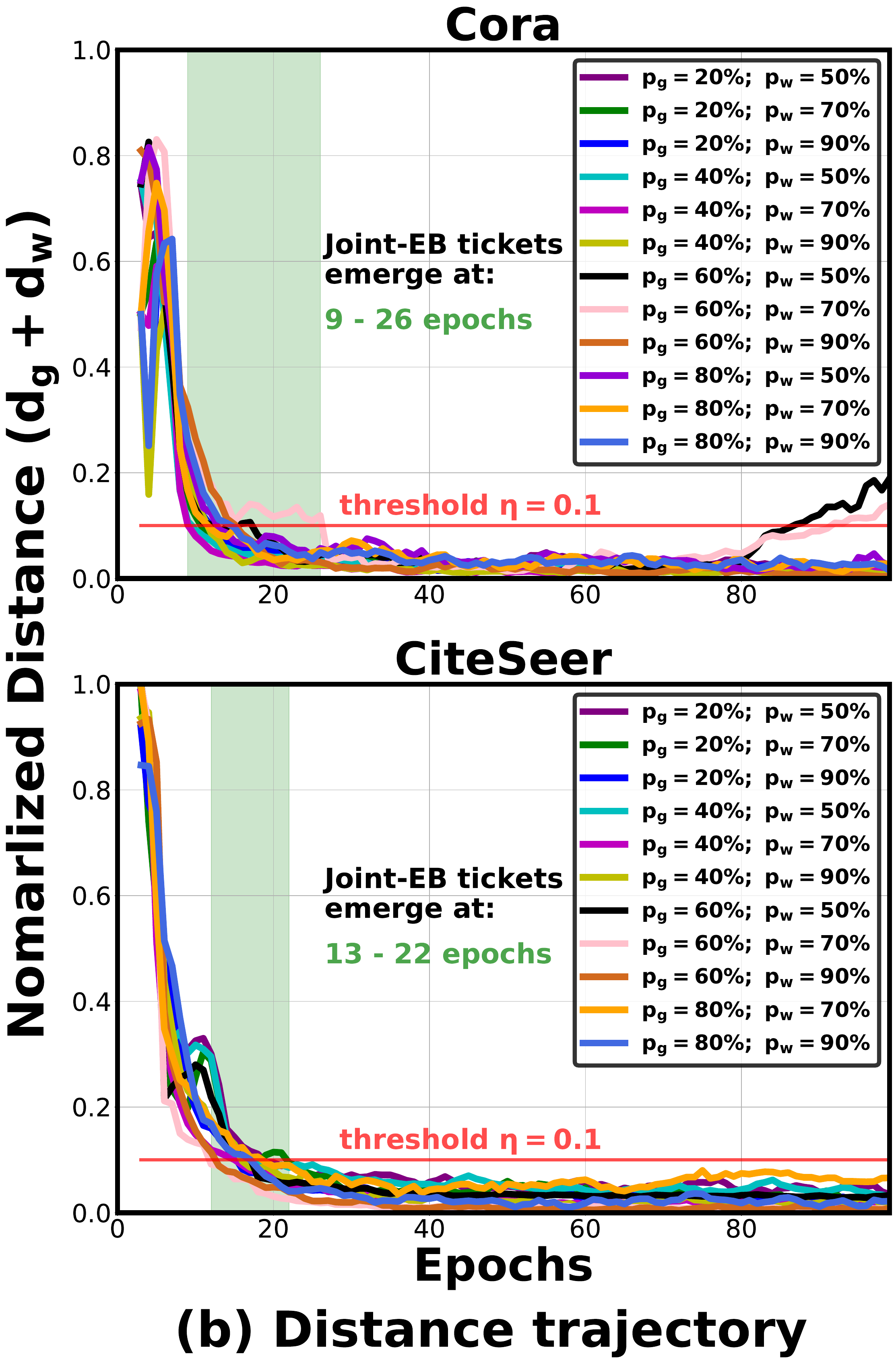}
     \vspace{-0.8em}
    \caption{(a) Retraining accuracy vs. epoch numbers at which both the subgraphs and subnetworks (i.e., joint-EB tickets) are drawn, for GCN networks \cite{kipf2017semi} on Cora and CiteSeer datasets, where $p_g$ indicates the graph pruning ratio and $p_w$ denotes the network pruning ratio, and (b) the distance's evolution along the training trajectories under different graph and network pruning ratio pairs.}
    \vspace{-1.5em}
    \label{fig:joint_EB_tickets}
\end{figure*}

\subsection{Finding 2: Joint-EB Tickets Exist}
\label{sec:joint_EB}
In this subsection, we first develop a co-sparsification framework to prune the GCN graphs and networks, and then 
show in a set of extensive experiments that joint-EB tickets exist across various models and datasets, and then propose a simple detector to detect the emergence of joint-EB tickets during co-sparsification of the GCN graphs and networks. 

\begin{figure*}[t]
    \centering
    \vspace{-0.5em}
    \includegraphics[width=0.9\linewidth]{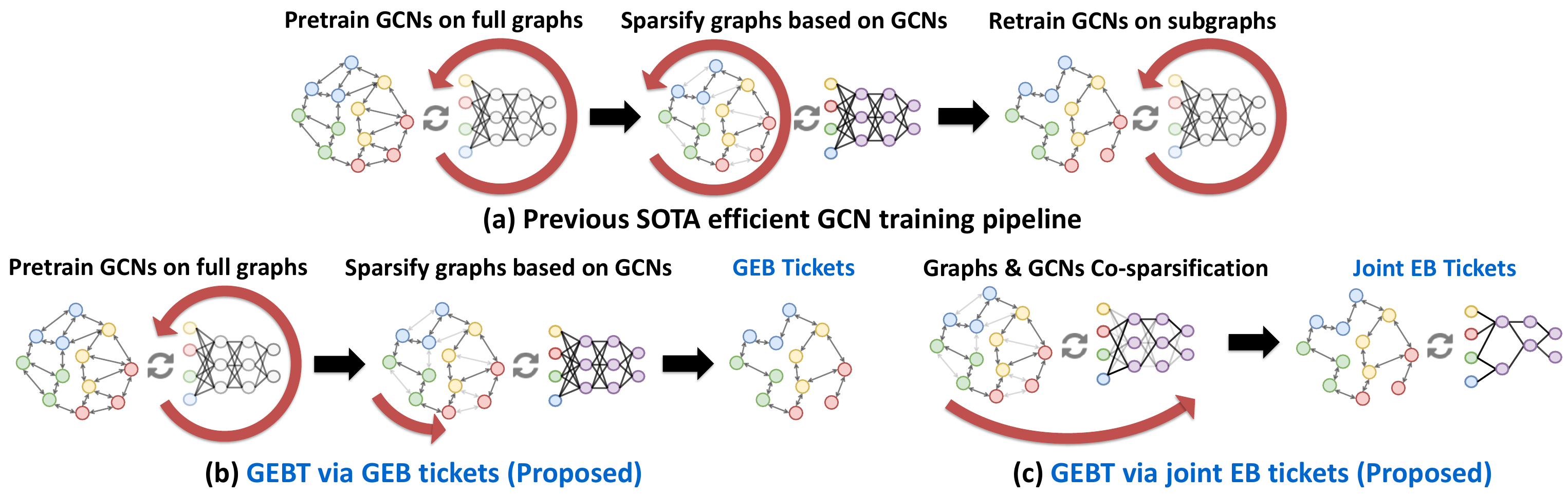}
    \vspace{-1em}
    \caption{An overview of the existing efficient GCN training pipeline and our GEBT training schemes via drawing GEB tickets and joint-EB tickets (red circle denotes the training process).}
    \vspace{-1em}
    \label{fig:overview}
\end{figure*}

\textbf{Co-sparsification of the GCN Graph and Network.} To explore the possibility of drawing joint-EB tickects between GCN graphs and networks, we first develop a co-sparsification framework, as described in Fig. \ref{fig:overview} (c) and Algorithm \ref{alg:joint_EB_detection}. Specifically, we iteratively update the GCN weights and graph adjacency matrices based on their corresponding loss functions formulated in Eq. (\ref{eq:gcn_loss}) and Eq. (\ref{eq:graph_loss}), respectively; after training for a certain epochs (e.g., 100 epochs), we then simultaneously prune the trained GCN graphs and networks using a magnitude-based pruning method \cite{han2015deep,frankle2018the}, and finally retrain the resulting pruned GCNs on the pruned graphs. Fig. \ref{fig:Co-optimization} shows the accuracy-FLOPS trade-offs of our co-sparsification framework when evaluating GCNs \cite{kipf2017semi} on Cora and CiteSeer graph datasets. We can see that co-sparsification can achieve up to 90\% sparsity in GCN weights while maintaining a comparable accuracy over the unpruned GCN graphs/networks.

\textbf{Existence of Joint-EB Tickets.} The existence of GEB tickets in GCN graphs and EB tickets in DNNs motivate our curiosity on the existence of joint-EB tickets between GCN graphs and networks. 
Fig. \ref{fig:joint_EB_tickets} (a) visualizes the retraining accuracies of the GCN subnetworks on subgraphs with both being drawn from different early epochs, which consistently indicates the existence of joint-EB tickets under an extensive set of experiments with different graph datasets, graph pruning ratios, and weight pruning ratios $\{G, p_g, p_w\}$. Furthermore, we can see that the joint-EB tickets emerge at the very early training stages (as early as $10$ epochs w.r.t. a total of 100 epochs), i.e., their retraining accuracy is comparable or even better than that of training the corresponding unpruned GCN graphs and networks or training the pruned graphs and unpruned GCN networks \cite{li2020sgcn}.

\textbf{Detection of Joint-EB Tickets.} 
We also develop a simple method to automatically detect the emergence of joint-EB tickets, of which the main idea is similar to the GEB tickets detector
but with an additional binary mask for drawing the GCN subnetwork. Similarly, in the binary masks, the pruned weights are set to 0 while the kept ones are set to 1, and the distance between subnetworks is characterize using the hamming distance between the corresponding binary masks following \cite{You2020Drawing} 
but we additionally define a binary mask of the drawn GCN subnetwork, where the pruned weights are 0 while the kept ones are 1. Therefore the distance between subnetworks is represented by the hamming distance between the corresponding binary masks following \cite{You2020Drawing}. For detecting the joint-EB tickets, we measure both the ``subgraph distance'' $d_g$ and ``subnetwork distance'' $d_w$ among consecutive epochs, resulting in three choices for the stop criteria (for a given the threshold $\eta$): 
(1) $d_g < \eta$;
(2) $d_w < \eta$;
(3) $d_g + d_w < \eta$.

Fig. \ref{fig:joint_EB_tickets} (b) leverages the third criterion to visualize the distance's trajectories of GCN networks on Cora and CiteSeer datasets, at different graph and network pruning ratio pairs $\{p_g, p_w\}$. 
\textbf{The ablation studies of all of the three criteria can be found in the Appendix.}
We can see that all criteria can effectively identify the emergence of joint-EB tickets, e.g., as early as 9 epochs w.r.t. a total of 100 epochs. 
Interestingly, the drawn joint-EB tickets can achieve a comparable or even better retraining accuracy than the subgraph and subnetwork pairs drawn at a later stages, which again implies the possibility of ``over-cooking'' as in the case of DNNs discussed in \cite{You2020Drawing}.
All results in this set of experiments consistently validate the existence of joint-EB tickets and the effectiveness of our joint-EB ticket detector.

\begin{figure}[t]
    \begin{minipage}{0.48\textwidth}
    \begingroup
    \removelatexerror
    \begin{algorithm}[H]
    \DontPrintSemicolon
    \KwIn{Graph $G = \{V, E, A, X\}$, 
    graph pruning ratio $p_g$, pretrained GCN weights $W$, and a FIFO queqe $Q$ with lenght $l$
    }
    
    \KwOut{The pruned adjacency matrix $A_p$}

   \While{$t$ (epoch) $< t_{max}$ }
   {
        GCN forward based on Eq. (\ref{eq:gcn})
        
        Update $A$ based on the $\mathcal{L}_{Graph}$ in Eq. (\ref{eq:graph_loss})
        
        Derive graph mask $m_t$ based on $A$ and ratio $p_g$
        
        Calculate the graph distance $d_g$ between $m_t$ and $m_{t-1}$ and add to $Q$
        
        \If{$\textbf{Max}(Q) < \eta$}
        {
            $t_{EB} = t$
            
            \textbf{Return} {$A_p = m_t \odot A$}
        }
   }
   
   
 
    \caption{GEB Tickets Identification} \label{alg:GEB_detection}
    \end{algorithm}
    \endgroup
\end{minipage}
\vspace{-2em}
\end{figure}

\begin{figure}[t]
    \begin{minipage}{0.48\textwidth}
    \begingroup
    \removelatexerror
    \begin{algorithm}[H]
    \DontPrintSemicolon
    \KwIn{Graph $G = \{V, E, A, X\}$,
    graph and weight pruning ratio $p_g$ and $p_w$, and a FIFO queqe $Q$ with lenght $l$
    }
    
    \KwOut{The pruned adjacency matrix $A_p$ and the pruned GCN weights $W_p$}
    
    Initialize the GCN weights $W$

   \While{$t$ (epoch) $< t_{max}$ }
   {
        GCN forward based on Eq. (\ref{eq:gcn})
        
        Update $W$ based on the $\mathcal{L}_{GCN}$ in Eq. (\ref{eq:gcn_loss})
        
        Update $A$ based on the $\mathcal{L}_{Graph}$ in Eq. (\ref{eq:graph_loss})
        
        Derive graph mask $m_t$ and network mask $n_t$ based on $A$, $W$ and pruning ratio $p_g$, $p_w$
        
        
        Calculate the distance $d_g$ between $m_t$ and $m_{t-1}$, $d_w$ between $n_t$ and $n_{t-1}$, and add $d_g + d_w$ to $Q$
        
        \If{$\textbf{Max}(Q) < \eta$}
        {
            $t_{EB} = t$
            
            \textbf{Return} $A_p = m_t \odot A$; $W_p = n_t \odot W$
        }
   }
   
   
 
    \caption{Joint-EB Tickets Identification} \label{alg:joint_EB_detection}
    \end{algorithm}
    \endgroup
\end{minipage}
\vspace{-2em}
\end{figure}

\vspace{-0.2em}
\subsection{Proposed GEBT:\! Efficient Training\! +\! Inference}
In this subsection, we present our proposed GEBT technique, which aims to leverage the existence of both GEB tickets and joint-EB tickets to develop a generic GCN efficient training framework. Note that GEBT achieves ``win-win": both efficient training and inference as the resulting trained GCN graphs and networks are naturally efficient. 
Here we will first describe the GEBT technique 
and then provide a complexity analysis to show GEBT's advantages.

\textbf{GEBT via GEB Tickets.}
Fig. \ref{fig:overview} (b) illustrates the overall pipeline of the proposed GEBT via drawing GEB tickets. Specifically, 
GEBT via drawing GEB tickets involves three steps: pretrain GCNs on the full graphs, train and sparsify the graph for identifying GEB tickets, and then retrain the GCN networks on the GEB tickets. The GEB ticket detection scheme is described in Algorithm \ref{alg:GEB_detection}.
Specifically, 
we use a magnitude-based pruning method \cite{han2015deep} to derive the graph mask (i.e., $m$) for calculating the graph distance between subgraphs from consecutive epochs and then store them into a first-in-first-out (FIFO) queue with a length of $l = 3$; The GEBT training will stop when the maximum graph distance is smaller than a specified threshold $\eta$ which is set to 0.1 in all our experiments, and return the GEB tickets (i.e., $A_p$) to be retrained.

\begin{figure*}[t]
    \centering
    \includegraphics[width=0.85\linewidth]{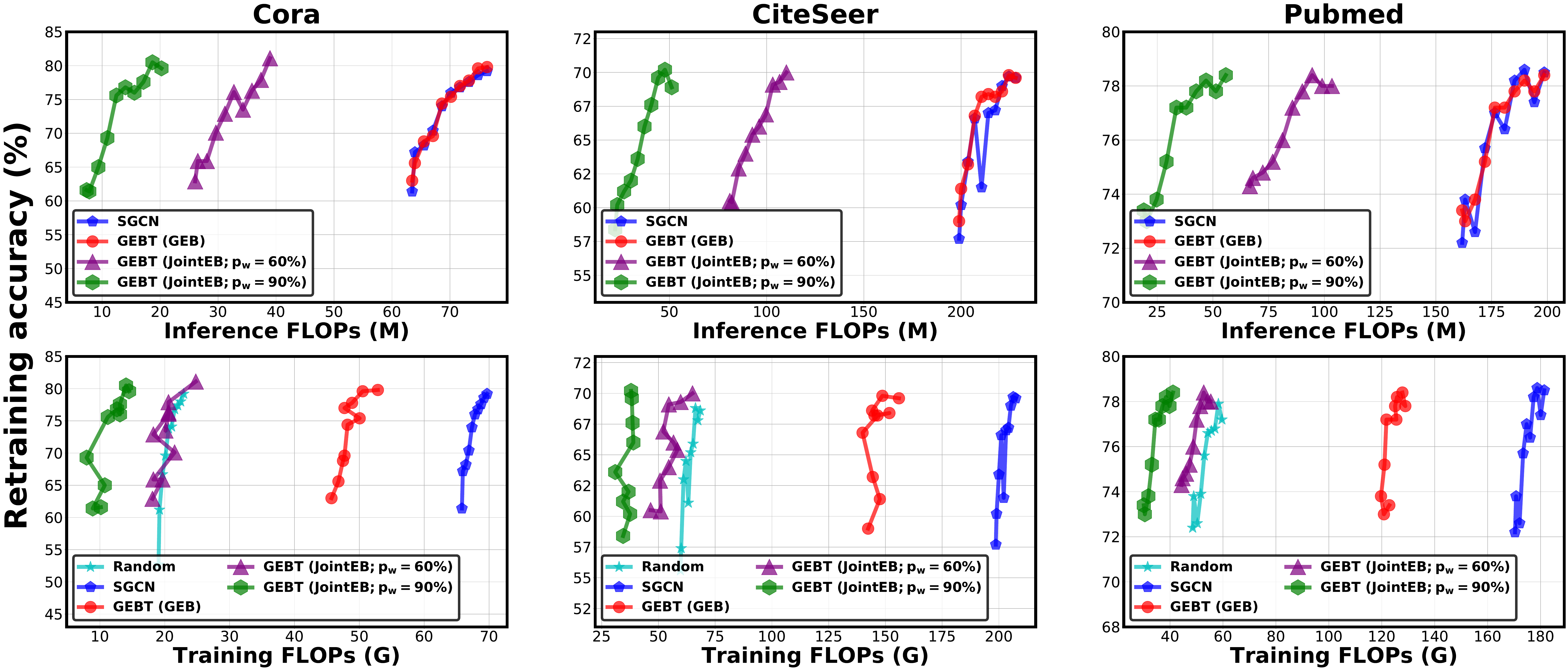}
    \vspace{-0.5em}
    \caption{Evaluating the retraining accuracy, training and inference FLOPs of the proposed GEBT over SOTA graph sparsification methods (Random pruning \cite{frankle2018the} and SGCN \cite{li2020sgcn}), under different graph and network sparsity pairs. Note that each method has a series of points for representing different graph sparsities ranging from 10\% to 90\%.}
    \vspace{-1em}
    \label{fig:overall_comparison_SP}
\end{figure*}

\textbf{GEBT via joint-EB Tickets.}
Fig. \ref{fig:overview} (c) shows the overall pipeline of the proposed GEBT technique via drawing joint-EB tickets. 
While SOTA efficient GCN training methods consist of three steps: (1) fully pretrain the GCN networks on the full graphs, (2) train and prune the graphs based on pretrained GCNs, and (3) retrain the GCN networks on pruned graph from scratch.
Accordingly, here GEBT via drawing joint-EB tickets only has two steps, it first follows the co-sparsification framework as described 
in previous sections
to prune and derive the GCN subgraph and subnetwork pairs, and then retrain the subnetwork on the drawn subgraph to restore accuracies. 
The joint-EB tickets detection scheme is described in Algorithm \ref{alg:joint_EB_detection}, where a FIFO queue is adopted for recording both the distance of subgraphs $d_g$ and subnetworks $d_w$ between consecutive epochs. GEBT training will stop when $d_g + d_w$ is smaller than a predefined threshold $\eta = 0.1$, and return the detected joint-EB tickets (i.e., $A_p$ and $W_p$) for further retraining.
Note that the initialization for retraining inherits from joint-EB tickets.


\textbf{Complexity Analysis of GEBT vs. SOTA Methods.} Here we provide complexity analysis to quantify the advantages of our GEBT technique. The time and memory complexity of GCN inferences can be captured by $\mathcal{O}(LMF + LNF^2)$ and $\mathcal{O}(LNF + LF^2)$, respectively, where 
$L, N, M$ and $F$ are the total number of GCN layers, nodes, edges, and features, respectively.
\cite{chiang2019cluster}. Assuming that drawing joint-EB tickets leads to $p_g$ and $p_w$ sparsities in GCN graphs and networks, respectively, then the inference time and memory complexity of GCNs resulting from our GEBT is $\mathcal{O}((1 - p_g)LMF + (1 - p_w)LNF^2)$ and $\mathcal{O}(LNF + (1 - p_w)LF^2)$, respectively.
Note that the corresponding training complexity will be scaled up by the total number of the required training epochs.

\vspace{-0.3em}
\section{Experiment Results}
\vspace{-0.1em}

\subsection{Experiment Setting}
\vspace{-0.1em}

\textbf{Models and Datasets.}
%
We evaluate the proposed methods over five representative GCN algorithms, i.e., GCN \cite{kipf2017semi}, GAT \cite{GAT}, GIN \cite{xu2018how}, GraphSAGE \cite{hamilton2017inductive}, and 7/14/28-layer deep ResGCNs \cite{li2020deepergcn}, on three citation graph datasets, i.e., Cora, CiteSeer, and Pubmed \cite{sen2008collective},
two inductive datasets, i.e., PPI and Reddit \cite{hamilton2017inductive},
and two large-scale graph datasets from \textit{Open Graph Benchmark (OGB)} \cite{hu2020ogb}, i.e., Ogbn-ArXiv for node classification and Ogbl-Collab for link prediction.
The statistics of these seven datasets are summarized in Tab. \ref{tab:dataset}.

\begin{table}[!b]
  \centering
  \vspace{-1.3em}
  \caption{The statistics of the adopted graph datasets.}
  \vspace{-1em}
  \resizebox{0.48\textwidth}{!}{
    \begin{tabular}{|l|c|c|c|c|c|}
    \hline
    \textbf{Dataset} & \textbf{Nodes} & \textbf{Edges} & \textbf{Features} & \textbf{Classes} & \textbf{Metric}  \\
    \hline
    \hline
    Cora & 2,708 & 5,429 & 1,433 & 7   & Accuracy \\
    \hline
    Citeseer & 3,312 & 4,732 & 3,703 & 6   & Accuracy \\
    \hline
    Pubmed & 19,717 & 44,338 & 500 & 3   & Accuracy \\
    \hline
    PPI & 56,944 & 818,716 & 50 & 121   & F1 Score \\
    \hline
    Ogbn-ArXiv & 169,343 & 1,166,243 & 128 & 40 & Accuracy  \\
    \hline
    Ogbl-Collab & 235,868 & 1,285,465 & 128 & 2 & Hits@50  \\
    \hline
    Reddit & 232,965 & 114,615,892 & 602 & 41 & F1 Score  \\
    \hline
    \end{tabular}%
    }
  \label{tab:dataset}%
\end{table}%

\textbf{Training Settings.}
We follow \cite{kipf2017semi} to train all the chosen two-layer GCN models on the three citation graph datasets and two inductive graph datasets, and follow \cite{li2020deepergcn} to train ResGCNs on OGB graphs. 
The detailed training settings are elaborated in the Appendix.


\begin{table}[t]
  \centering
  \caption{GEBT vs. SOTA GCN methods on citation graphs, where $\uparrow$ and $\downarrow$ denote the improvement over original models.
}
  \vspace{-0.8em}
  \resizebox{0.48\textwidth}{!}{
\begin{tabular}{|l|ccc|ccc|}
\hline
 \multicolumn{1}{|l|}{\multirow{2}{*}{\textbf{Methods}}} & \multicolumn{3}{c|}{\textbf{Accuracy (\%)}} & \multicolumn{3}{c|}{\textbf{Inference FLOPs (M)}} \\
\cline{2-7}  & \multicolumn{1}{c}{\textbf{Cora}} & \multicolumn{1}{c}{\textbf{CiteSeer}} & \multicolumn{1}{c|}{\textbf{Pumbed}} &
\multicolumn{1}{c}{\textbf{Cora}} & \multicolumn{1}{c}{\textbf{CiteSeer}} & \multicolumn{1}{c|}{\textbf{Pumbed}} \\
\hline
\hline
GCN & 80.9  & 69.4  & 79.0  &  77.95  & 231.6  & 203.1  \\
GraphSAGE & 82.5  & 71.0  & 78.9  & 6239  & 19654  & 15868  \\
GAT & 82.1  & 72.1  & 79.0  &  623.6 & 1853  & 1624 \\
GIN & 81.6  & 70.9  & 79.1  &  77.95  & 231.6  & 203.1 \\
\hline
\textbf{GEBT (GCN)} & 81.1 ($\uparrow$0.2) & 70.5 ($\uparrow$1.1) & 78.5 ($\downarrow$0.5)  & 24.9 ($\uparrow$3.1$\times$) & 51.2 ($\uparrow$4.5$\times$) & 55.8 ($\uparrow$3.6$\times$) \\
\textbf{GEBT (GraphSAGE)} & 82.6 ($\uparrow$0.1) & 70.7 ($\downarrow$0.3) & 78.0 ($\downarrow$0.9)  & 624 ($\uparrow$10$\times$)  &  1965 ($\uparrow$10$\times$)  &  4760 ($\uparrow$3.3$\times$) \\
\textbf{GEBT (GAT)} & 82.2 ($\uparrow$0.1) & 74.1 ($\uparrow$2.0) & 79.8 ($\uparrow$0.8) &  149 ($\uparrow$4.2$\times$) &  382 ($\uparrow$4.9$\times$)  &  446 ($\uparrow$3.6$\times$) \\
\textbf{GEBT (GIN)} & 82.4 ($\uparrow$0.8) & 71.4 ($\uparrow$0.5) & 79.7 ($\uparrow$0.6) &  20.2 ($\uparrow$3.8$\times$) &  90.5 ($\uparrow$2.6$\times$)  &  55.8 ($\uparrow$3.6$\times$) \\
\hline
\textbf{Overall Improv.} &  \multicolumn{3}{c|}{\textbf{$\downarrow$0.9 $\sim$ $\uparrow$2.0}} & \multicolumn{3}{c|}{\textbf{$\uparrow$2.6$\times$  $\sim$ $\uparrow$10.0$\times$}} \\
\hline
\end{tabular}%
}
\label{tab:compare_GCNs}%
\end{table}%

%

\textbf{Baselines and Evaluation Metrics.}
We evaluate the effectiveness of the proposed GEBT's improved training and inference efficiency in terms of the node classification accuracy (or F1 Score, Hits@50), inference FLOPs, and total training FLOPs, as compared to other graph sparsifiers, i.e., random pruning \cite{frankle2018the} and SGCN \cite{li2020sgcn}, and \textbf{ten standard SOTA GCN algorithms} using unpruned graphs. 

\begin{table}[t]
  \centering
  \caption{GEBT vs. SOTA efficient GCN methods on PPI.
   }
  \vspace{-1em}
  \resizebox{\linewidth}{!}{
    \begin{tabular}{|l|c|c|c|}
        \hline
        \multirow{2}[4]{*}{\textbf{Methods}} & \multicolumn{3}{c|}{\textbf{PPI (56K nodes and 818K edges)}} \\
        \cline{2-4}  & \multicolumn{1}{c|}{\textbf{F1 Scores (\%)}} & \multicolumn{1}{c|}{\textbf{Infer. FLOPs (G)}} & \multicolumn{1}{c|}{\textbf{Train. FLOPs (T)}} \\
        \hline
        \hline
        GAT & 98.2  & 3.15  & 18.9 \\
        ResGCN & 98.5  & 47.85  & 287.1 \\
        ClusterGCN & 99.3  & 35.0  & 210.0 \\
        \hline
        GraphSAGE & 61.2  & 155.8  & 934.8 \\
        VRGCN & 97.8  & 76.75  & 460.5 \\
        GraphSAINT & 98.1  & 35.0  & 210.0 \\
        L2-GCN & 96.8  & 35.0  & 210.0 \\
        N-GCN & 65.0  & 30.42  & 182.5 \\
        \hline
        \textbf{GEBT (GAT) vs. GAT} & 98.8 ($\uparrow$0.6)  & 1.84 ($\uparrow$1.7$\times$)  & 11.2 ($\uparrow$1.7$\times$) \\
        \textbf{GEBT (ResGCN) vs. ResGCN} & 98.6 ($\uparrow$0.1)  & 24.15 ($\uparrow$2.0$\times$)  & 147.8 ($\uparrow$1.9$\times$) \\
        \textbf{GEBT (ClusterGCN) vs. ClusterGCN} & 99.2 ($\downarrow$0.1)  & 19.31 ($\uparrow$1.8$\times$)   &  118.2 ($\uparrow$1.8$\times$)  \\
        \hline
        \textbf{Overall Improv.} & \textbf{$\downarrow$0.1 $\sim$ $\uparrow$38}  & \textbf{$\uparrow$1.7$\times$ $\sim$ $\uparrow$84.1$\times$}  &  \textbf{$\uparrow$1.7$\times$ $\sim$ $\uparrow$ 83.5$\times$} \\
        \hline
    \end{tabular}%
  }
  \vspace{-1.5em}
\label{tab:ppi}%
\end{table}

\vspace{-0.3em}
\subsection{GEBT over SOTA Sparsifiers}
We compare the proposed GEBT with existing SOTA GCN sparsification pipelines \cite{li2020sgcn} on the three citation graphs to evaluate the effectiveness of GEBT.
Fig. \ref{fig:overall_comparison_SP} shows that GEBT consistently outperforms all competitors in terms of measured accuracies and computational costs (i.e., training and inference FLOPs) trade-offs.
Specifically, GEBT via GEB tickets achieves 24.7\%$\sim$32.1\% training FLOPs reduction while offering comparable accuracies ($\downarrow$1.4\%$\sim\uparrow$4.9\%) across a wide range of graph pruning ratios, as compared to SGCN.
Furthermore, GEBT via joint-EB tickets even aggressively reaches 80.2\%$\sim$85.6\% and 84.6\%$\sim$87.5\% reduction in training FLOPs and inference FLOPs, respectively, over SGCN when pruning the GCN networks up to 90\% sparsity, meanwhile leading to a comparable accuracy range ($\downarrow$1.3\%$\sim\uparrow$1.4\%).
This set of experiments verify (1) the efficiency benefits of the GEBT framework and (2) the high-quality of the drawn GEB tickets and joint-EB tickets.


\begin{table}[t]
  \centering
  \caption{GEBT vs. SOTA efficient GCN methods on Reddit.
  }
  \vspace{-1em}
  \resizebox{\linewidth}{!}{
    \begin{tabular}{|l|c|c|c|}
        \hline
        \multirow{2}[4]{*}{\textbf{Methods}} & \multicolumn{3}{c|}{\textbf{Reddit (232K nodes and 11M edges)}} \\
        \cline{2-4}  & \multicolumn{1}{c|}{\textbf{F1 Scores (\%)}} & \multicolumn{1}{c|}{\textbf{Infer. FLOPs (G)}} & \multicolumn{1}{c|}{\textbf{Train. FLOPs (T)}} \\
        \hline
        \hline
        GCN & 95.6  & 52.3  & 470.9 \\
        GraphSAGE & 95.4  & 2396.7  &  21570.7 \\
        \hline
        FastGCN & 93.7  & 958.7  & 8628.3 \\
        VRGCN & 96.3  & 956.6  & 8609.7 \\
        ClusterGCN & 96.6  & 226.8  & 2041.1 \\
        GraphSAINT & 96.6  & 226.8  & 2041.1 \\
        GTTF (GraphSAGE) & 95.9  & 2396.7  & 21570.7 \\
        L2-GCN & 94.0  & 226.8  & 2041.1 \\
        \hline
        \textbf{GEBT (GCN) vs. GCN} & 95.8 ($\uparrow$0.2)  & 29.3 ($\uparrow$1.8$\times$)  & 266.9 ($\uparrow$1.7$\times$) \\
        \textbf{GEBT (GraphSAGE) vs. GraphSAGE} & 97.1 ($\uparrow$1.7)  & 1198.4 ($\uparrow$2.0$\times$) & 10929.1 ($\uparrow$2.0$\times$) \\
        \hline
        \textbf{Overall Improv.} &  \textbf{$\uparrow$0.5 $\sim$ $\uparrow$3.4} &  \textbf{$\uparrow$1.8$\times$ $\sim$ $\uparrow$81.8$\times$}   &  \textbf{$\uparrow$1.7$\times$ $\sim$ $\uparrow$80.8$\times$} \\
        \hline
    \end{tabular}%
  }
  \vspace{-.5em}
\label{tab:reddit}%
\end{table}

\subsection{GEBT over SOTA GCNs}
To evaluate the benefits of GEBT,
we first compare the performance of GEBT over four SOTA GCN algorithms on three citation graphs. As shown in Tab. \ref{tab:compare_GCNs}, GEBT consistently outperforms all the baselines in terms of efficiency-accuracy trade-offs.
Specifically, GEBT achieves 2.6$\times$ $\sim$ 10$\times$ inference FLOPs reduction,
while offering a comparable accuracy ($\downarrow$0.9\% $\sim$ $\uparrow$2.0\%), as compared to SOTA GCN algorithms.
We further evaluate GEBT with eight SOTA methods on two large datasets, PPI and Reddit, and show the comparisons in Tables \ref{tab:ppi} and \ref{tab:reddit}, respectively, where ($\uparrow$) and ($\downarrow$) denote improvement over the \textit{original} models, and ``Overall Improv.'' denotes the best improvement over all SOTA baselines. 
GEBT again consistently achieves the best efficiency-accuracy trade-offs, e.g., reducing inference FLOPs (up to 84.1\%) and training FLOPs (up to 83.5\%) under comparable or even higher F1-micro scores ($\downarrow$0.1\% $\sim$ $\uparrow$38\%).

\vspace{-0.4em}
\section{Conclusion}
\vspace{-0.1em}

GCNs have gained increasing attention thanks to their SOTA performance on graphs. However, the notorious challenge of GCN training and inference limits their application to large real-world graphs and hinders the exploration of deeper and more sophisticated GCN graphs. To this end, we advocate graph-network co-optimization and explore the possibility of drawing early-bird tickets when sparsifying GCN graphs.
Specifically, we for the first time discover the existence of GEB tickets that emerge at the very early stage when sparsifying GCN graphs, and propose a simple yet effective detector to automatically identify their emergence. Furthermore, we develop a generic efficient GCN training framework dubbed GEBT that can significantly boost the efficiency of GCN training and inference by enabling co-sparsification and drawing joint-EB of GCNs. Experiments on various GCN models and datasets consistently validate our GEB finding and the effectiveness of our GEBT. 

\section*{Acknowledgements}
We would like to acknowledge the funding support from the NSF EPCN program (Award ID: 1934767) for this project.

\bibliography{aaai22}

\begin{thebibliography}{33}
\providecommand{\natexlab}[1]{#1}

\bibitem[{Canziani, Paszke, and Culurciello(2016)}]{canziani2016analysis}
Canziani, A.; Paszke, A.; and Culurciello, E. 2016.
\newblock An analysis of deep neural network models for practical applications.
\newblock \emph{arXiv preprint arXiv:1605.07678}.

\bibitem[{Chen et~al.(2021{\natexlab{a}})Chen, Sui, Chen, Zhang, and
  Wang}]{chen2021unified}
Chen, T.; Sui, Y.; Chen, X.; Zhang, A.; and Wang, Z. 2021{\natexlab{a}}.
\newblock A unified lottery ticket hypothesis for graph neural networks.
\newblock In \emph{International Conference on Machine Learning}, 1695--1706.
  PMLR.

\bibitem[{Chen et~al.(2021{\natexlab{b}})Chen, Cheng, Wang, Gan, Wang, and
  Liu}]{chen2021earlybert}
Chen, X.; Cheng, Y.; Wang, S.; Gan, Z.; Wang, Z.; and Liu, J.
  2021{\natexlab{b}}.
\newblock EarlyBERT: Efficient BERT Training via Early-bird Lottery Tickets.

\bibitem[{Chiang et~al.(2019)Chiang, Liu, Si, Li, Bengio, and
  Hsieh}]{chiang2019cluster}
Chiang, W.-L.; Liu, X.; Si, S.; Li, Y.; Bengio, S.; and Hsieh, C.-J. 2019.
\newblock Cluster-GCN: An efficient algorithm for training deep and large graph
  convolutional networks.
\newblock In \emph{Proceedings of the 25th ACM SIGKDD International Conference
  on Knowledge Discovery \& Data Mining}, 257--266.

\bibitem[{Frankle and Carbin(2019)}]{frankle2018the}
Frankle, J.; and Carbin, M. 2019.
\newblock The Lottery Ticket Hypothesis: Finding Sparse, Trainable Neural
  Networks.
\newblock In \emph{International Conference on Learning Representations}.

\bibitem[{Gao et~al.(2019)Gao, Yang, Zhang, Zhou, and Hu}]{gao2019graphnas}
Gao, Y.; Yang, H.; Zhang, P.; Zhou, C.; and Hu, Y. 2019.
\newblock Graphnas: Graph neural architecture search with reinforcement
  learning.
\newblock \emph{arXiv preprint arXiv:1904.09981}.

\bibitem[{Geng et~al.(2020)Geng, Li, Shi, Wu, Wang, Li, Haghi, Tumeo, Che,
  Reinhardt et~al.}]{geng2020awb}
Geng, T.; Li, A.; Shi, R.; Wu, C.; Wang, T.; Li, Y.; Haghi, P.; Tumeo, A.; Che,
  S.; Reinhardt, S.; et~al. 2020.
\newblock AWB-GCN: A graph convolutional network accelerator with runtime
  workload rebalancing.
\newblock In \emph{53rd IEEE/ACM Int. Symp. Microarchit.(MICRO)}, 1--15.

\bibitem[{Geng et~al.(2021)Geng, Wu, Zhang, Tan, Xie, You, Herbordt, Lin, and
  Li}]{10.1145/3466752.3480113}
Geng, T.; Wu, C.; Zhang, Y.; Tan, C.; Xie, C.; You, H.; Herbordt, M.; Lin, Y.;
  and Li, A. 2021.
\newblock \emph{I-GCN: A Graph Convolutional Network Accelerator with Runtime
  Locality Enhancement through Islandization}, 1051–1063.
\newblock New York, NY, USA: Association for Computing Machinery.
\newblock ISBN 9781450385572.

\bibitem[{Hamilton, Ying, and Leskovec(2017)}]{hamilton2017inductive}
Hamilton, W.; Ying, Z.; and Leskovec, J. 2017.
\newblock Inductive representation learning on large graphs.
\newblock In \emph{Advances in neural information processing systems},
  1024--1034.

\bibitem[{Han, Mao, and Dally(2015)}]{han2015deep}
Han, S.; Mao, H.; and Dally, W.~J. 2015.
\newblock Deep compression: Compressing deep neural networks with pruning,
  trained quantization and huffman coding.
\newblock \emph{arXiv preprint arXiv:1510.00149}.

\bibitem[{Hu et~al.(2020)Hu, Fey, Zitnik, Dong, Ren, Liu, Catasta, and
  Leskovec}]{hu2020ogb}
Hu, W.; Fey, M.; Zitnik, M.; Dong, Y.; Ren, H.; Liu, B.; Catasta, M.; and
  Leskovec, J. 2020.
\newblock Open graph benchmark: Datasets for machine learning on graphs.
\newblock \emph{arXiv preprint arXiv:2005.00687}.

\bibitem[{Kersting et~al.(2016)Kersting, Kriege, Morris, Mutzel, and
  Neumann}]{KKMMN2016}
Kersting, K.; Kriege, N.~M.; Morris, C.; Mutzel, P.; and Neumann, M. 2016.
\newblock Benchmark Data Sets for Graph Kernels.
\newblock \url{http://graphkernels.cs.tu-dortmund.de}.

\bibitem[{Kipf and Welling(2016)}]{kipf2016semi}
Kipf, T.~N.; and Welling, M. 2016.
\newblock Semi-supervised classification with graph convolutional networks.
\newblock \emph{arXiv preprint arXiv:1609.02907}.

\bibitem[{Kipf and Welling(2017)}]{kipf2017semi}
Kipf, T.~N.; and Welling, M. 2017.
\newblock Semi-Supervised Classification with Graph Convolutional Networks.
\newblock In \emph{International Conference on Learning Representations
  (ICLR)}.

\bibitem[{Li et~al.(2020{\natexlab{a}})Li, Xiong, Thabet, and
  Ghanem}]{li2020deepergcn}
Li, G.; Xiong, C.; Thabet, A.; and Ghanem, B. 2020{\natexlab{a}}.
\newblock Deepergcn: All you need to train deeper gcns.
\newblock \emph{arXiv preprint arXiv:2006.07739}.

\bibitem[{Li et~al.(2020{\natexlab{b}})Li, Zhang, Tian, Jin, Fardad, and
  Zafarani}]{li2020sgcn}
Li, J.; Zhang, T.; Tian, H.; Jin, S.; Fardad, M.; and Zafarani, R.
  2020{\natexlab{b}}.
\newblock SGCN: A Graph Sparsifier Based on Graph Convolutional Networks.
\newblock In \emph{Pacific-Asia Conference on Knowledge Discovery and Data
  Mining}, 275--287. Springer.

\bibitem[{Liu et~al.(2018)Liu, Sun, Zhou, Huang, and
  Darrell}]{liu2018rethinking}
Liu, Z.; Sun, M.; Zhou, T.; Huang, G.; and Darrell, T. 2018.
\newblock Rethinking the value of network pruning.
\newblock \emph{arXiv preprint arXiv:1810.05270}.

\bibitem[{Mukund~Kalibhat, Balaji, and Feizi(2020)}]{mukund2020winning}
Mukund~Kalibhat, N.; Balaji, Y.; and Feizi, S. 2020.
\newblock Winning Lottery Tickets in Deep Generative Models.
\newblock \emph{arXiv e-prints}, arXiv--2010.

\bibitem[{Peng et~al.(2020)Peng, Hong, Chen, and Zhao}]{peng2020learning}
Peng, W.; Hong, X.; Chen, H.; and Zhao, G. 2020.
\newblock Learning Graph Convolutional Network for Skeleton-Based Human Action
  Recognition by Neural Searching.
\newblock In \emph{AAAI}, 2669--2676.

\bibitem[{Sen et~al.(2008)Sen, Namata, Bilgic, Getoor, Galligher, and
  Eliassi-Rad}]{sen2008collective}
Sen, P.; Namata, G.; Bilgic, M.; Getoor, L.; Galligher, B.; and Eliassi-Rad, T.
  2008.
\newblock Collective classification in network data.
\newblock \emph{AI magazine}, 29(3): 93--93.

\bibitem[{Simonovsky and Komodakis(2017)}]{simonovsky2017dynamic}
Simonovsky, M.; and Komodakis, N. 2017.
\newblock Dynamic edge-conditioned filters in convolutional neural networks on
  graphs.
\newblock In \emph{Proceedings of the IEEE conference on computer vision and
  pattern recognition}, 3693--3702.

\bibitem[{Tailor, Fernandez-Marques, and Lane(2020)}]{tailor2020degree}
Tailor, S.~A.; Fernandez-Marques, J.; and Lane, N.~D. 2020.
\newblock Degree-Quant: Quantization-Aware Training for Graph Neural Networks.
\newblock \emph{arXiv preprint arXiv:2008.05000}.

\bibitem[{Veličković et~al.(2018)Veličković, Cucurull, Casanova, Romero,
  Liò, and Bengio}]{GAT}
Veličković, P.; Cucurull, G.; Casanova, A.; Romero, A.; Liò, P.; and Bengio,
  Y. 2018.
\newblock Graph Attention Networks.
\newblock In \emph{International Conference on Learning Representations}.

\bibitem[{Wu et~al.(2020)Wu, Pan, Chen, Long, Zhang, and
  Philip}]{wu2020comprehensive}
Wu, Z.; Pan, S.; Chen, F.; Long, G.; Zhang, C.; and Philip, S.~Y. 2020.
\newblock A comprehensive survey on graph neural networks.
\newblock \emph{IEEE Transactions on Neural Networks and Learning Systems}.

\bibitem[{Xu et~al.(2018)Xu, Hu, Leskovec, and Jegelka}]{xu2018powerful}
Xu, K.; Hu, W.; Leskovec, J.; and Jegelka, S. 2018.
\newblock How powerful are graph neural networks?
\newblock \emph{arXiv preprint arXiv:1810.00826}.

\bibitem[{Xu et~al.(2019)Xu, Hu, Leskovec, and Jegelka}]{xu2018how}
Xu, K.; Hu, W.; Leskovec, J.; and Jegelka, S. 2019.
\newblock How Powerful are Graph Neural Networks?
\newblock In \emph{International Conference on Learning Representations}.

\bibitem[{Yan et~al.(2020)Yan, Deng, Hu, Liang, Feng, Ye, Zhang, Fan, and
  Xie}]{yan2020hygcn}
Yan, M.; Deng, L.; Hu, X.; Liang, L.; Feng, Y.; Ye, X.; Zhang, Z.; Fan, D.; and
  Xie, Y. 2020.
\newblock Hygcn: A gcn accelerator with hybrid architecture.
\newblock In \emph{2020 IEEE International Symposium on High Performance
  Computer Architecture (HPCA)}, 15--29. IEEE.

\bibitem[{Ying et~al.(2018)Ying, You, Morris, Ren, Hamilton, and
  Leskovec}]{ying2018hierarchical}
Ying, Z.; You, J.; Morris, C.; Ren, X.; Hamilton, W.; and Leskovec, J. 2018.
\newblock Hierarchical graph representation learning with differentiable
  pooling.
\newblock In \emph{Advances in neural information processing systems},
  4800--4810.

\bibitem[{You et~al.(2020)You, Li, Xu, Fu, Wang, Chen, Baraniuk, Wang, and
  Lin}]{You2020Drawing}
You, H.; Li, C.; Xu, P.; Fu, Y.; Wang, Y.; Chen, X.; Baraniuk, R.~G.; Wang, Z.;
  and Lin, Y. 2020.
\newblock Drawing Early-Bird Tickets: Toward More Efficient Training of Deep
  Networks.
\newblock In \emph{International Conference on Learning Representations}.

\bibitem[{Zeng et~al.(2019)Zeng, Zhou, Srivastava, Kannan, and
  Prasanna}]{zeng2019accurate}
Zeng, H.; Zhou, H.; Srivastava, A.; Kannan, R.; and Prasanna, V. 2019.
\newblock Accurate, efficient and scalable graph embedding.
\newblock In \emph{2019 IEEE International Parallel and Distributed Processing
  Symposium (IPDPS)}, 462--471. IEEE.

\bibitem[{Zhang et~al.(2018)Zhang, Cui, Neumann, and Chen}]{zhang2018end}
Zhang, M.; Cui, Z.; Neumann, M.; and Chen, Y. 2018.
\newblock An end-to-end deep learning architecture for graph classification.
\newblock In \emph{Thirty-Second AAAI Conference on Artificial Intelligence}.

\bibitem[{Zhang et~al.(2021)Zhang, You, Fu, Geng, Li, and
  Lin}]{DBLP:journals/corr/abs-2109-08983}
Zhang, Y.; You, H.; Fu, Y.; Geng, T.; Li, A.; and Lin, Y. 2021.
\newblock G-CoS: GNN-Accelerator Co-Search Towards Both Better Accuracy and
  Efficiency.
\newblock \emph{CoRR}, abs/2109.08983.

\bibitem[{Zheng et~al.(2020)Zheng, Zong, Cheng, Song, Ni, Yu, Chen, and
  Wang}]{zheng2020robust}
Zheng, C.; Zong, B.; Cheng, W.; Song, D.; Ni, J.; Yu, W.; Chen, H.; and Wang,
  W. 2020.
\newblock Robust Graph Representation Learning via Neural Sparsification.
\newblock In \emph{Proceedings of the 37th International Conference on Machine
  Learning}, volume 119, 11458--11468. PMLR.

\end{thebibliography}


\end{document}